\documentclass[conference]{IEEEtran}
\IEEEoverridecommandlockouts
\usepackage{cite}
\usepackage{amsmath,amssymb,amsfonts}
\usepackage{algorithmic}
\usepackage{graphicx}
\usepackage{textcomp}
\usepackage{xcolor}

\usepackage{hyperref}

\usepackage{booktabs}
\usepackage{bbm}
\usepackage{multicol}
\usepackage{multirow}
\usepackage{colortbl}
\definecolor{mygray}{gray}{.9}
\usepackage{bbding}

\def\BibTeX{{\rm B\kern-.05em{\sc i\kern-.025em b}\kern-.08em
    T\kern-.1667em\lower.7ex\hbox{E}\kern-.125emX}}
\begin{document}

\title{
{Prohibited Items Segmentation via Occlusion-aware Bilayer Modeling}
\thanks{\IEEEauthorrefmark{1} Corresponding author.}
}


\author{
    \IEEEauthorblockN{Yunhan Ren\textsuperscript{1,2}, Ruihuang Li\textsuperscript{1}, Lingbo Liu\textsuperscript{2}, Changwen Chen\textsuperscript{1,*}}
    \IEEEauthorblockA{
        \textsuperscript{1}Department of Computing, The Hong Kong Polytechnic University, Hong Kong, China \\
        \textsuperscript{2}Research Institute of Multiple Agents and Embodied Intelligence, Pengcheng Laboratory, Shenzhen, China \\
        yunhan.ren@connect.polyu.hk, csrhli@comp.polyu.edu.hk, liulingbo918@gmail.com, changwen.chen@polyu.edu.hk
    }
}

\maketitle

\begin{abstract}
Instance segmentation of prohibited items in security X-ray images is a critical yet challenging task. This is mainly caused by the significant appearance gap between prohibited items in X-ray images and natural objects, as well as the severe overlapping among objects in X-ray images. To address these issues, we propose an occlusion-aware instance segmentation pipeline designed to identify prohibited items in X-ray images. Specifically, to bridge the representation gap, we integrate the Segment Anything Model (SAM) into our pipeline, taking advantage of its rich priors and zero-shot generalization capabilities. To address the overlap between prohibited items, we design an occlusion-aware bilayer mask decoder module that explicitly models the occlusion relationships. To supervise occlusion estimation, we manually annotated occlusion areas of prohibited items in two large-scale X-ray image segmentation datasets, PIDray and PIXray. We then reorganized these additional annotations together with the original information as two occlusion-annotated datasets, PIDray-A and PIXray-A. Extensive experimental results on these occlusion-annotated datasets demonstrate the effectiveness of our proposed method. The datasets and codes are available at: \href{https://github.com/Ryh1218/Occ}{https://github.com/Ryh1218/Occ}.
\end{abstract}


\section{Introduction} \label{sec:intro}
Security inspection is a critical process in various real-world contexts, such as airports and train stations \cite{Intro1, Intro2}. Typically, human inspectors are responsible for examining scanned X-ray images generated by security inspection machines to identify potentially prohibited items. With the advancement of deep learning technologies, the computer vision community has made efforts to achieve automatic object detection and segmentation by applying general instance segmentation models \cite{BNT, FCOS, SAPN}.

\begin{figure}[t]
\centering
\includegraphics[width=0.48\textwidth]{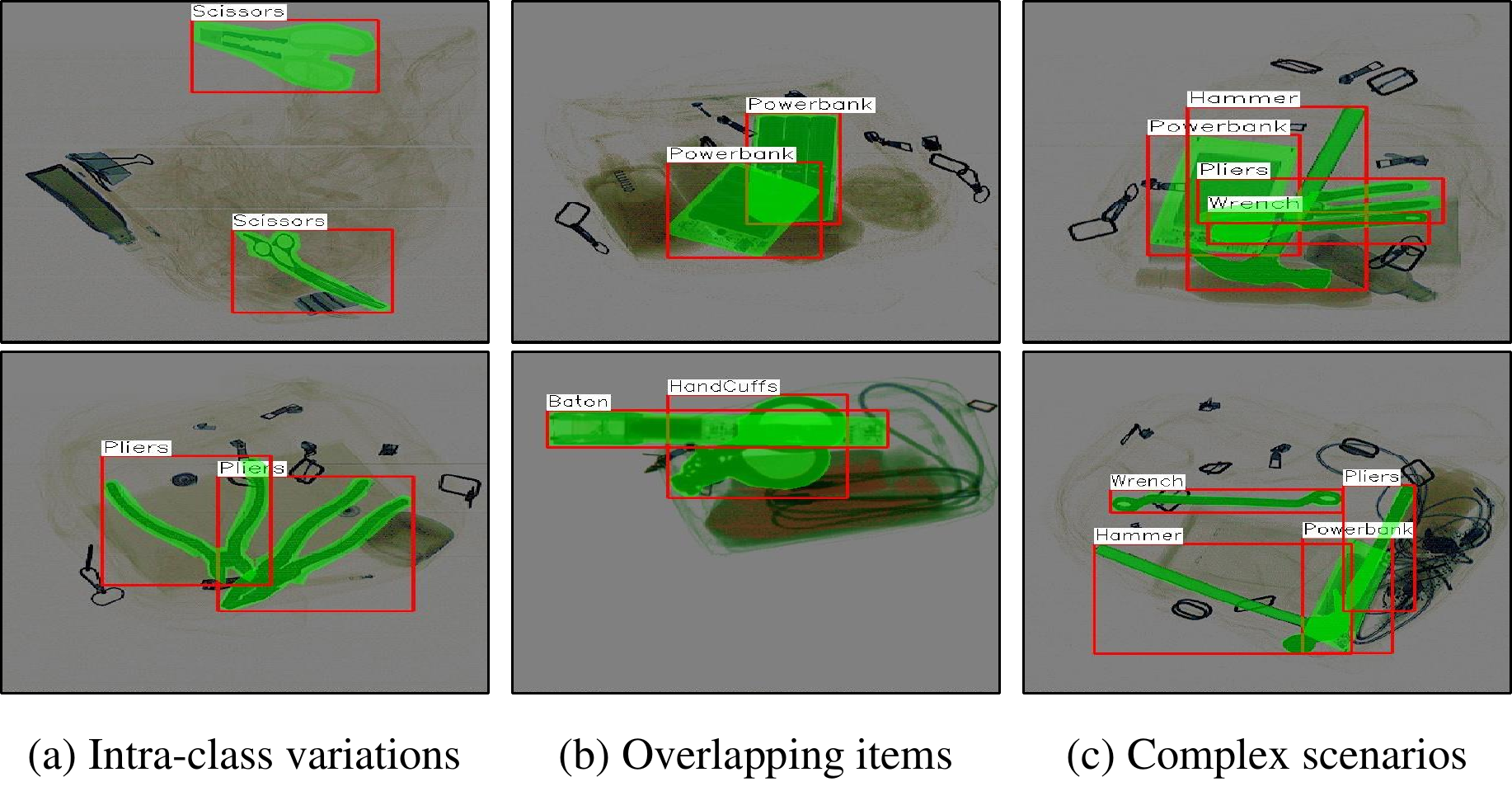}
\vspace{-2mm}
\caption{Unique characteristics of X-ray images where multiple prohibited items with intra-class variations overlap with each other, causing difficulties in distinguishing and segmenting them.}
\vspace{-5mm}
\label{ximage}
\end{figure}

However, these methods, primarily designed for natural images, face two main challenges when applied to X-ray images: (1) Objects in X-ray images exhibit a significant appearance gap compared to those in natural images, which is primarily caused by different materials absorbing X-rays to varying degrees. (2) Objects in X-ray images often overlap with other items while showing substantial intra-class variations, as illustrated in Fig. \ref{ximage}. These complexities challenge traditional pipelines in understanding the semantic meanings and spatial relationships of prohibited items within X-ray images. Therefore, deep learning frameworks with strong generalization capabilities and specialized techniques for handling occlusion are essential for accurately detecting and segmenting prohibited items in X-ray images \cite{BCNet, BCNetA}.

\begin{figure*}[!t]
\centering
\includegraphics[width=0.9\textwidth]{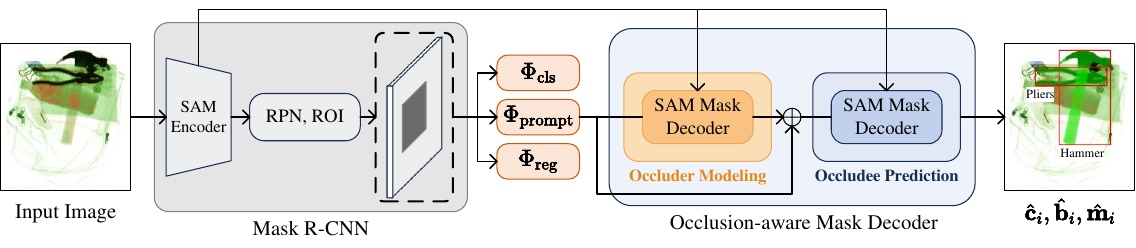}
\vspace{-2mm}
\caption{Demonstration of the proposed pipeline. A frozen SAM Encoder extracts features from input images. Then, RPN and ROI Align predict the normalized ROI features $\hat{\mathbf{v}}_{i}$. $\Phi_{\text{cls}}$, $\Phi_{\text{reg}}$ and $\Phi_{\text{prompt}}$ represents the semantic head, localization head, and prompt head. They obtain category predictions, IoU predictions, and sparse prompt embeddings from ROI features, respectively. The proposed occlusion-aware bilayer mask decoder then takes prompt embeddings and image embeddings to segment the target objects with the guidance of their occluding objects.}
\vspace{-3mm}
\label{pipeline}
\end{figure*}
To this end, we propose an occlusion-aware instance segmentation pipeline specifically designed for segmenting prohibited items in X-ray images. To precisely capture the general representation of X-ray images and fully adapt to the appearance patterns of prohibited items, we incorporate the Segment Anything Model (SAM) into our pipeline, leveraging its rich image priors and powerful generalization capabilities \cite{SAM}. Recent advances in applying SAM across various research fields have demonstrated its effectiveness in object recognition and generalization. For example, recent work by \cite{RSP} employs prompt learning methods to harness SAM's capabilities for instance segmentation tasks in remote sensing applications. Additionally, studies have explored SAM's application in medical image segmentation, showing promising results despite certain limitations when compared to specialized medical segmentation models \cite{MedSAM}. Furthermore, integrating SAM with other advanced techniques, such as generative adversarial networks, has been proposed to enhance its performance on pose estimation \cite{SAM6}. These ongoing research efforts provide a solid foundation for adopting SAM to address specific challenges.

Specifically, inspired by \cite{RSP}, our pipeline employs a frozen SAM Image Encoder as its backbone to generate precise image representations. It then converts the Region of Interest (ROI) features generated by Mask R-CNN into prompts and encodes them using the SAM Prompt Encoder. The SAM Mask Decoder is utilized to regress the mask predictions based on the image representations and prompts. To minimize the influence of overlap between prohibited items, inspired by \cite{BCNetA, BCNet}, we design an occlusion-aware bilayer mask decoder module that explicitly distinguishes overlapping areas among prohibited items. Specifically, we employ two sequential SAM Mask Decoders with separate training objectives: the first decoder estimates the occlusion area for each target object, while the second regresses the binary mask of the target item guided by the predicted occlusion area. The intermediate optimized image embeddings and prompt embeddings from the first decoder serve as guidance for explicitly modeling occlusion relationships. To effectively train the first decoder, corresponding ground-truth annotations of each occlusion area are required, while none of the existing prohibited instance segmentation X-ray datasets contain such information. Therefore, we introduce two large-scale occlusion-annotated datasets, PIDray-A and PIXray-A, derived from the large-scale X-ray image segmentation datasets PIDray and PIXray \cite{PIDray, PIDrayOld, PIXray}. For each prohibited item, we manually annotate the overlapping areas of other prohibited objects and record the segmentation masks of these areas as additional annotation information. Consequently, the proposed datasets maintain the same number of images as their baseline versions but include additional occlusion annotations to supervise the occlusion-aware mask decoder. To demonstrate the effectiveness of our proposed pipeline, we conduct extensive experiments on the two occlusion-annotated datasets. The experimental results indicate the superiority of our pipeline compared to the state-of-the-art segmentation methods.

In summary, this paper will contribute in the following ways: 
\begin{itemize}
    \item We propose an occlusion-aware instance segmentation pipeline designed for automatic security inspection.
    \item We design an occlusion-aware bilayer mask decoder module with a guidance mechanism to effectively handle occlusion between prohibited items. 
    \item We create two large-scale security datasets with extra occlusion annotations to meet the need for ground-truth annotations of occlusion areas.
\end{itemize}

\section{Related Work} \label{RelatedWork}
\subsection{Instance Segmentation}
Instance segmentation is concerned with locating objects within images and generating a semantic mask for each individual object \cite{Survey}. Two-stage instance segmentation methods typically consist of a region proposal phase followed by a classification and refinement process. Among two-stage methods, the Mask R-CNN series \cite{MR-CNN} serve as robust baselines for subsequent advancements. This method first detects the bounding boxes of target objects and then performs semantic segmentation within each ROI area \cite{BCNet}. To enhance feature interaction, PANet \cite{PANet} integrates a bottom-up pathway grounded in Feature Pyramid Network (FPN) \cite{FPN}. Meanwhile, Cascade R-CNN \cite{Cascade} employs a sequential architecture of detection heads with different IoUs (Intersection over Union) to generate high-precision bounding boxes. Expanding the cascading strategy, Hybrid Task Cascade \cite{HTC} introduces a multitask, multistage hybrid structure that enriches contextual information, leading to superior performance.

\subsection{Occlusion Handling} \label{occlusionhandling}
Traditional occlusion handling methods typically involve sophisticated algorithms that exploit symmetry and consistency in stereo matching \cite{OHTra1} or incorporate contextual information into random field models for object recognition and segmentation \cite{OHTra2}. Advanced methods also include predicting occlusion overlap order by building a scoring histogram \cite{OHA1} or incorporating top-down category-specific reasoning and shape prediction into an energy minimization framework \cite{OHA2}. Additionally, \cite{BCNet, BCNetA} explicitly model occluding and occluded objects separately, using graph convolutional networks to process ROI features of occluding objects and then guide the segmentation of target occluded objects.

\subsection{Prohibited Item Segmentation Benchmarks} \label{prohibitedbenchmark}
Recently, several datasets have been proposed to facilitate research on prohibited item detection. \cite{DS1} and \cite{DS2} contain prohibited items with complicated backgrounds and overlapping, but the numbers of images and prohibited items are insufficient. Similarly, \cite{SIXray} introduces a large-scale security inspection dataset named SIXray, but only a small proportion of images (0.84\%) contain prohibited items. In response to the need for prohibited item segmentation in real-world scenarios, \cite{PIDray} and \cite{PIDrayOld} present a large-scale benchmark, PIDray, consisting of 124,486 X-ray images, with 47,677 of them containing prohibited items across 12 categories. Additionally, the PIXray dataset \cite{PIXray} is proposed with 15 classes of 15,201 prohibited items, which are all annotated with instance-level masks. These datasets provide robust experimental references for our research.

\section{Methodology}
\subsection{Model Architecture}
Our proposed framework is based on Mask R-CNN \cite{MR-CNN}, incorporating SAM Image Encoder \cite{SAM} as the backbone, along with its prompt encoder to transform ROI features generated by Mask R-CNN into SAM-style prompt embedding. An occlusion-aware mask decoder module is proposed for decoding the prompted image features to semantic masks while distinguishing the occlusion relationships between prohibited items. Fig. \ref{pipeline} demonstrates our proposed pipeline. Formally, consider a training dataset $\mathcal{D}_{\text{train}} = {(\mathbf{x}_{1}, \mathcal{Y}_{1}), (\mathbf{x}_{2}, \mathcal{Y}_{2}), \dots, (\mathbf{x}_{N}, \mathcal{Y}_{N}) }$, where $\mathbf{x}_{i} \in \mathbb{R}^{H \times W \times 3}$ represents an image while $\mathcal{Y}_{i}= \{(\mathbf{b}_{i}, \mathbf{c}_{i}, \mathbf{m}_{i}, \mathbf{m}^{e}_{i}) \}$ represents corresponding ground-truth information set. Specifically, $\mathbf{b}_{i} \in \mathbb{R}^{n_{i}\times 4}$ stands for $n_{i}$ bounding box annotations of prohibited items in image $\mathbf{x}_{i}$, $\mathbf{c}_{i} \in \mathbb{R}^{n_{i}\times C}$ represents the corresponding set of semantic labels, $\mathbf{m}_{i} \in \mathbb{R}^{n_{i}\times H \times W}$ stands for semantic mask of each prohibited item, and $\mathbf{m}^{e}_{i}\in \mathbb{R}^{n_{i}\times H \times W}$ represents the part of the semantic masks of other annotated object which overlapping the area of $\mathbf{b}_{i}$, if it has any. In our pipeline, for an input image $\mathbf{x}_{i}$, its image representation feature $\mathbf{f}^{\text{img}}_{i}$, predicted regional proposals $\hat{\mathbf{o}}_{i}$, and normalized ROI features $\hat{\mathbf{v}}_{i}$ are defined as:
\begin{equation}
\begin{aligned}
\mathbf{f}^{\text{img}}_{i} &= \Phi_{\text{enc}}(\mathbf{x}_{i}) \\
\hat{\mathbf{o}}_{i} &= \Phi_{\text{rpn}}(\mathbf{f}^{\text{img}}_{i}) \\
\hat{\mathbf{v}}_{i} &= \Phi_{\text{roi}}(\mathbf{f}^{\text{img}}_{i}+\text{PE}, \hat{\mathbf{o}}_{i}) \\
\end{aligned}
\end{equation}
, where $\Phi_{\text{enc}}$ is a frozen SAM image encoder, $\text{PE}$ is positional encoding to retain spatial information, $\Phi_{\text{rpn}}$ and $\Phi_{\text{roi}}$ are standard RPN and ROI align in Mask R-CNN pipeline.

Predicted ROI features $\hat{\mathbf{v}}_{i}$ are then processed by the semantic head $\Phi_{\text{cls}}$, the localization head $\Phi_{\text{reg}}$, and the prompt head $\Phi_{\text{prompt}}$ to produce the category prediction $\hat{\mathbf{c}}_{i}$, the bounding box prediction $\hat{\mathbf{b}}_{i}$ and the SAM-format prompt embedding $\mathbf{f}^{\text{sparse}}_{i}$. 

To accurately estimate the target object mask predictions $\hat{\mathbf{m}}_{i}$ while avoiding being influenced by its occluding object mask $\hat{\mathbf{m}}_{i}^{e}$, we introduce an occlusion-aware bilayer mask decoder architecture, inspired by \cite{BCNet}, which is designed to explicitly model the relationships between the occluding object (occluder) and the target object (occludee). The occlusion-aware mask decoder $\Phi_{\text{m}}$ explicitly models the relationships between $\hat{\mathbf{m}}_{i}$ and $\hat{\mathbf{m}}_{i}^{e}$ (if any exist):
\begin{equation}
\begin{aligned}
\{ \hat{\mathbf{c}}_{i}, \hat{\mathbf{b}}_{i}, \mathbf{f}^{\text{sparse}}_{i} \} &= \{ \Phi_{\text{cls}}(\hat{\mathbf{v}}_{i}), \Phi_{\text{reg}}(\hat{\mathbf{v}}_{i}), \Phi_{\text{prompt}}(\hat{\mathbf{v}}_{i})\} \\
\hat{\mathbf{m}}_{i}, \hat{\mathbf{m}}_{i}^{e} &= \Phi_{\text{m}}(\mathbf{f}^{\text{img}}_{i}, \mathbf{f}^{\text{sparse}}_{i})
\end{aligned}
\end{equation}

\begin{figure}[!t]
    \centering
    \includegraphics[width=0.9\linewidth]{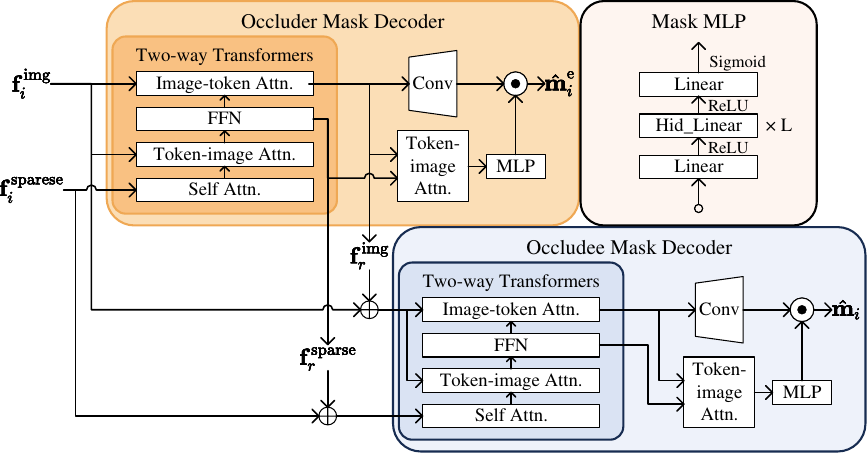}
    \caption{A detailed implementation of the bilayer occlusion-aware mask decoder structure. Two mask decoders separately predict the masks of the occluder and occludee, providing information about occlusion relationships as one of the learning objectives for the pipeline.}
    \vspace{-3mm}
    \label{fig:decoder}
\end{figure}

\subsection{Bilayer Mask Decoder}
\subsubsection{Module structure}
Fig. \ref{fig:decoder} illustrates the structure of our proposed bilayer mask decoder, which consists of two sequentially connected SAM mask decoders. The first decoder predicts the mask of the area where the occluder overlaps the occludee, and its output guides the subsequent decoder in generating the occludee's prediction.

\subsubsection{The occluder decoder}
The occluder decoder $\Phi_{r}$ aims to explicitly model the occlusion between the occluder and occludee using the SAM mask decoder. Formally, it takes the sparse prompt embedding $\mathbf{f}_{i}^{\text{sparse}}$ and image embedding $\mathbf{f}_{i}^{\text{img}}$ as inputs to produce the overlapping area prediction of the occluder mask, denoted as $\hat{\mathbf{m}}_{i}^{e}$. Simultaneously, benefiting from the SAM mask decoder's involvement, $\mathbf{f}_{i}^{\text{sparse}}$ and $\mathbf{f}_{i}^{\text{img}}$ are optimized in terms of global information, with their optimized versions denoted as $\mathbf{f}_{r}^{\text{sparse}}$ and $\mathbf{f}_{r}^{\text{img}}$.

\subsubsection{The residual guidance mechanism}
Unlike BCNet, which uses only simple GCN-optimized ROI features for guidance, we incorporate information from both image and prompt embeddings. Specifically, we element-wise add these transformer-optimized embeddings ($\mathbf{f}_{r}^{\text{sparse}}$ and $\mathbf{f}_{r}^{\text{img}}$) to the original embeddings separately before feeding them into the occludee mask decoder for mask regression. This residual guidance mechanism ensures the occlusion area is explicitly emphasized through cross-attention, helping the model learn occlusion relationships and thereby improving overall performance.

\subsubsection{The occludee decoder}
Finally, the occludee decoder $\Phi_{e}$ employs another SAM mask decoder to generate the final occludee prediction, using the refined embeddings $\mathbf{f}_{e}^{\text{img}}$ and $\mathbf{f}_{e}^{\text{sparse}}$ as inputs. The above process can be expressed as follows:
\begin{equation}
    \begin{aligned}
        \hat{\mathbf{m}}_{i}^{e}, \mathbf{f}_{r}^{\text{img}}, \mathbf{f}_{r}^{\text{sparse}} &= \Phi_{r}(\text{Cat}(\mathbf{t}_{\text{mask}}, \mathbf{t}_{\text{IoU}}, \mathbf{f}^{\text{sparse}}_{i}), \mathbf{f}^{\text{img}}_{i}) \\
        \mathbf{f}_{e}^{\text{img}}, \mathbf{f}_{e}^{\text{sparse}} &= \mathbf{f}_{i}^{\text{img}} + \mathbf{f}_{r}^{\text{img}}, \mathbf{f}_{i}^{\text{sparse}} + \mathbf{f}_{r}^{\text{sparse}} \\
        \hat{\mathbf{m}}_{i} &= \Phi_{e}(\text{Cat}(\mathbf{t}_{\text{mask}}, \mathbf{t}_{\text{IoU}}, \mathbf{f}^{\text{sparse}}_{e}), \mathbf{f}_{e}^{\text{img}})\\
    \end{aligned}
\end{equation}
, where $\mathbf{t}_{\text{mask}}$ and $ \mathbf{t}_{\text{IoU}}$ are learnable tokens and are concatenated with prompt embeddings, $\text{Cat}(\cdot)$ represents concatenate operation. To ensure the accuracy of the occluder mask decoder during training, its intermediate occlusion area mask prediction $\hat{\mathbf{m}}_{i}^{e}$ is used to compare with ground-truth annotations as an extra learning objective.

\subsection{End-to-end Parameter Learning}
Based on the Mask R-CNN architecture \cite{MR-CNN}, instance segmentation can be formed as an end-to-end parameter learning problem. This process is guided by a multi-task loss function $\mathcal{L}$, which simultaneously optimizes the RPN's region proposal loss $\mathcal{L}_{\text{rpn}}$ and the ROI's object recognition loss. The ROI object recognition loss has three components: classification loss $\mathcal{L}_{\text{cls}}$, bounding box regression loss $\mathcal{L}_{\text{reg}}$, and mask prediction loss $\mathcal{L}_{\text{seg}}$. We adopt $\mathcal{L}_{\text{rpn}}$, $\mathcal{L}_{\text{cls}}$ and $\mathcal{L}_{\text{reg}}$ from \cite{MR-CNN}.

In our framework, we modify the $\mathcal{L}_{\text{seg}}$ with extra occluder mask loss. The original mask prediction loss, denoted as $\mathcal{L}_{\text{pred}}$, is typically computed using the Binary Cross Entropy loss, defined as:
\begin{equation}
\begin{split}
\mathcal{L}_{\text{pred}}(\hat{\mathbf{m}}_{i}, \mathbf{m}_{i}) = -\frac{1}{w \cdot h} \sum_{j=1}^{w} \sum_{k=1}^{h} [ \mathbf{m}_{i,j,k} \log(\hat{\mathbf{m}}_{i,j,k}) \\
+ (1 - \mathbf{m}_{i,j,k}) \log(1 - \hat{\mathbf{m}}_{i,j,k})]
\end{split}
\end{equation}
, where $\hat{\mathbf{m}}_{i}$ and $\mathbf{m}_{i}$ are the predicted mask and the ground-truth mask, $w$ and $h$ are the width and height of the mask, respectively, and the loss is taken over all pixels $(j,k)$ in the mask.

In our proposed pipeline, an additional objective loss is implemented to let both the occludee mask prediction $\hat{\mathbf{m}}_{i}$ and occluder mask prediction $\hat{\mathbf{m}}_{i}^{e}$ contribute to the overall mask prediction loss. Formally, the segmentation loss $\mathcal{L}_{\text{seg}}$ is defined as:
\begin{equation}
\mathcal{L}_{\text{seg}} = \mathcal{L}_{\text{pred}}(\hat{\mathbf{m}}_{i}, \mathbf{m}_{i}) + \lambda \cdot \mathcal{L}_{\text{pred}}(\hat{\mathbf{m}}_{i}^{e}, \mathbf{m}_{i}^{e})
\end{equation}
, where $\lambda$ is a hyperparameter that balances the contributions of the occluder and occludee losses.

The total loss is the weighted sum of the above losses:
\begin{equation}
\mathcal{L} = \frac{1}{M}\sum_{i}^M \mathcal{L}_{rpn}^i + \frac{1}{N} \sum_{j}^N (\mathcal{L}_{cls}^j + \mathbbm{1}^j(\mathcal{L}_{reg}^j + \mathcal{L}_{seg}^j))
\end{equation}
, where $\mathbbm{1}$ represents the indicator function that is used to validate positive matches.

\begin{table}[t]
\centering
    \caption{Statistical details of the PIDray-A and PIXray-A datasets}
    \begin{tabular}[b]{c c c ccc c c}
        \toprule
            \multirow{4}{*}{Type} &
            \multirow{4}{*}{Count} &
            \multicolumn{4}{c}{PIDray-A} &
            \multicolumn{2}{c}{PIXray-A} \\
            \cmidrule(lr){3-6}\cmidrule(lr){7-8}
            & & \multirow{2.5}{*}{Train} &
            \multicolumn{3}{c}{Test} &
            \multirow{2.5}{*}{Train} &
            \multirow{2.5}{*}{Test} \\
            \cmidrule(lr){4-6}
            & & & Easy & Hard & Hide \\
        \midrule
            \multirow{4}{*}{Images} & Total & 76913 & 24758 & 9746 & 13069  & 3560 & 1486 \\
            & Anno & 29454 & 9482 & 3733 & 5005 & 3560 & 1486 \\
            & Multi & 7411 & 0 & 3733 & 3 & 2891 & 1193 \\
            & Occlu & 4785 & 0 & 2434 & 0 & 1550 & 654 \\
        \midrule
            \multirow{2}{*}{Annos} & Total & 39708 & 9482 & 8892 & 5008 & 10709 & 4508 \\
            & Extra & 11080 & 0 & 5585 & 0 & 4114 & 1765 \\ 
        \bottomrule
    \end{tabular}
    \vspace{-5mm}
    \label{Datasets}
\end{table}%

\begin{table*}[!t]
    \centering
    \caption{Experimental results of object detection and instance segmentation on the PIDray-A and PIXray-A datasets}
    \begin{tabular}[b]{c ccc ccc ccc ccc}
        \toprule
            \multirow{2.5}{*}{Model} & \multicolumn{6}{c}{PIDray-A} & \multicolumn{6}{c}{PIXray-A} \\
            \cmidrule(lr){2-7} \cmidrule(lr){8-13}
            & $AP_{b}$ & $AP_{b}^{50}$ & $AP_{b}^{75}$ & $AP_{m}$ & $AP_{m}^{50}$ & $AP_{m}^{75}$ & $AP_{b}$ & $AP_{b}^{50}$ & $AP_{b}^{75}$ & $AP_{m}$ & $AP_{m}^{50}$ & $AP_{m}^{75}$\\
        \midrule
            Mask R-CNN & 58.7 & 76.2 & 67.2 & 50.2 & 73.4 & 58.2 & 65.2 & 91.1 & 75.3 & 56.1 & 86.7 & 61.7 \\
            Cascade MR-CNN & 65.0 & 78.1 & 72.1 & 53.6 & 76.0 & 62.4 & 71.8 & 91.2 & 81.0 & 57.1 & 86.2 & 63.1 \\
            PIDrayNet & \textcolor{blue}{\textbf{66.6}} & 81.7 & \textcolor{blue}{\textbf{74.3}} & 53.4 & 78.6 & 61.4 & \textcolor{red}{\textbf{72.4}} & 91.6 & 81.3 & \textcolor{blue}{\textbf{57.7}} & 87.1 & 64.1 \\
            RSPrompter & 66.0 & \textcolor{blue}{\textbf{83.1}} & \textcolor{blue}{\textbf{74.3}} & \textcolor{blue}{\textbf{56.5}} & \textcolor{blue}{\textbf{81.4}} & \textcolor{blue}{\textbf{64.4}} & \textcolor{blue}{\textbf{72.0}} & \textcolor{blue}{\textbf{94.3}} & \textcolor{blue}{\textbf{83.1}} & \textcolor{red}{\textbf{58.9}} & \textcolor{blue}{\textbf{90.1}} & \textcolor{red}{\textbf{66.4}} \\
            \rowcolor{mygray}
            Our Model & \textcolor{red}{\textbf{67.5}} & \textcolor{red}{\textbf{84.3}} & \textcolor{red}{\textbf{75.8}} & \textcolor{red}{\textbf{57.6}} & \textcolor{red}{\textbf{82.6}} & \textcolor{red}{\textbf{65.8}} & \textcolor{blue}{\textbf{72.0}} & \textcolor{red}{\textbf{94.4}} & \textcolor{red}{\textbf{83.7}} & \textcolor{red}{\textbf{58.9}} & \textcolor{red}{\textbf{90.6}} & \textcolor{blue}{\textbf{66.2}} \\
        \bottomrule
    \end{tabular}
    \label{tab:mainresults}
\end{table*}%


\section{Experiments}
\subsection{Datasets}
We primarily evaluate our instance segmentation model on the proposed occlusion-annotated PIDray-A and PIXray-A datasets, which are modified based on the large-scale X-ray image datasets PIDray and PIXray. The PIDray dataset \cite{PIDrayOld, PIDray} consists of 124,486 X-ray images, including 47,677 images that contain prohibited items in 12 categories. The test set of PIDray is additionally divided into three subsets: the easy subset contains images with only one exposed prohibited item, the hard subset includes images with multiple prohibited items, and the hidden subset contains images with one prohibited object that is deliberately concealed to test the model's ability under extreme circumstances. As for the PIXray dataset \cite{PIXray}, it contains 5,046 X-ray images, all of which have one or more prohibited items. In total, the PIXray dataset includes 15,201 prohibited items with bounding boxes and semantic mask annotations across 15 object categories. In our experiment, we use an 8:2 train-test split for PIXray.

\begin{figure}[t]
\centering
\includegraphics[width=0.48\textwidth]{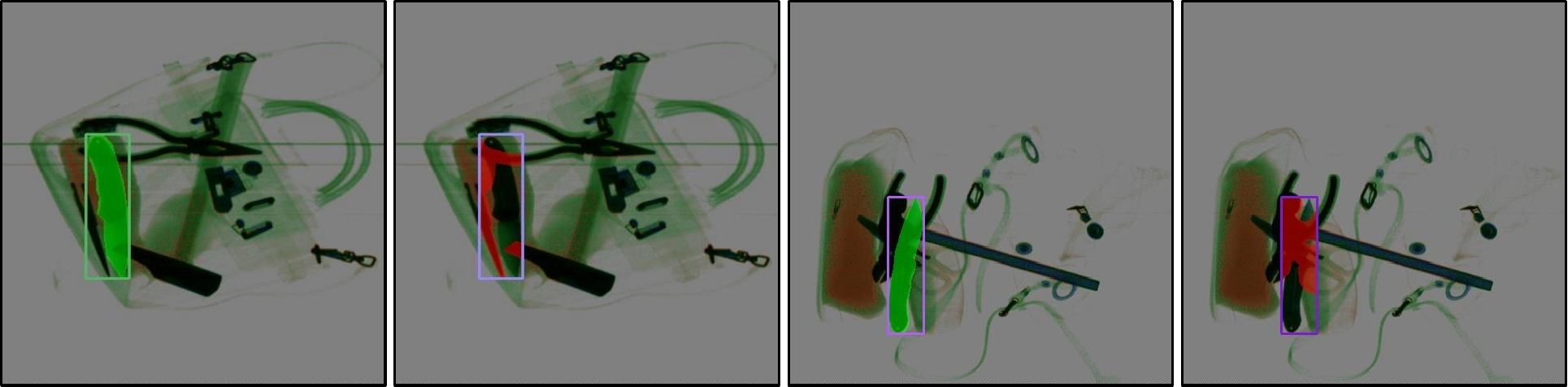}
\caption{Occlusion annotations in the proposed PIDray-A dataset. Green masks denote the target object (occludee) mask annotations, while red masks represent the background object (occluder) annotations for each occludee.}
\vspace{-3mm}
\label{oimage}
\end{figure}

To optimize the occluder mask decoder, ground-truth occluder annotations are necessary for calculating $\mathcal{L}_{\text{seg}}$. Consequently, we add additional occlusion annotations for contraband items in PIDray and PIXray, reorganizing them into two occlusion-annotated datasets. Specifically, we first identify items whose masks have at least 5\% of their area covered by other objects' bounding boxes and mark them as occludees. Then, we verify whether these occludees' masks intersect with their occluders' masks, recording the intersecting mask area as occlusion annotations. For cases without intersections, we discard them as non-occluding instances. Fig. \ref{oimage} demonstrates the example visualization of occlusion annotations on representative X-ray images, indicating the extra emphasis of our pipeline on occluding areas. Statistic details of the proposed PIDray-A and PIXray-A datasets are presented in Tab. \ref{Datasets}. Specifically, the 'Anno,' 'Multi,' and 'Occlu' rows indicate whether the images contain prohibited items, multiple prohibited items, or occluded items, respectively. The 'Extra' row indicates the number of prohibited items with additional occlusion annotations. It can be concluded that more than half of images with more than one prohibited item suffer from overlapping of prohibited items, indicating that an occlusion-aware instance segmentation pipeline is highly demanded.

\subsection{Implementation details}
In line with the commonly used COCO metrics, we employ mean average precision (mAP) to evaluate our model's performance. Specifically, we focus on mAP at three IoU thresholds: $T = 0.5$, $T = 0.75$, and $\{T\} = (0.50:0.05:0.95)$, for both bounding boxes and masks, denoted as $AP_{b}$ and $AP_{m}$. All input images are resized to a resolution of $512 \times 512$ to generate visual patches for the SAM Image Encoder. To balance computational cost and model performance, we adopt ViT-Large as the backbone of SAM. The hyperparameter $\lambda$ for segmentation loss calculation is set to 0.25. For optimization, we use the AdamW optimizer with an initial learning rate of $1e^{-6}$. After a warm-up phase of 50 iterations, during which the learning rate linearly increases to $1e^{-4}$, we apply a Cosine Annealing scheduler \cite{CAS} to gradually decay it back to $1e^{-6}$ over the training period, ensuring stable training. To augment the training data, we employ horizontal flipping and random jittering, implemented via the MMDetection framework. All experiments are conducted on a single NVIDIA A800 Tensor Core GPU with 80GB VRAM, using a batch size of 64. We train the model for 50 epochs on the PIDray-A dataset and 60 epochs on the PIXray-A dataset.

\subsection{Experiment Results}
We quantitatively compare our proposed methods with other state-of-the-art methods. Specifically, we compare our models with Mask R-CNN \cite{MR-CNN}, Cascade Mask R-CNN \cite{Cascade}, SDANet \cite{PIDrayOld}, and the PIDray baseline \cite{PIDray} (namely PIDrayNet). Moreover, we run RSPrompter \cite{RSP} with the same backbone to indicate the impact of the proposed occlusion-aware bilayer mask decoder. The results of these comparisons are detailed in Tab. \ref{tab:mainresults}, where our proposed model achieves the optimal results on 10 out of 12 metrics while obtaining suboptimal results on the other 2 metrics. Specifically, on the PIDray-A dataset, our proposed model achieves 0.9\% improvements compared with suboptimal results on both bounding box mAP and semantic mask mAP. Regarding the PIXray-A dataset, our model obtains either optimal or suboptimal results. These results demonstrate the effectiveness of our proposed model on automatic instance segmentation of prohibited items.

We also demonstrate the results on subsets of the PIDray-A dataset, as shown in Tab. \ref{tab:subsetresults}. Our proposed model performs the best results on 4 out of 6 subsets. However, our model shows no clear advantage in detecting and segmenting hidden objects. We assume that this is primarily because the PIDrayNet uses specialized X-ray image optimizing tricks for hidden object filtering, while our model relies on a simple RPN for detection.

\begin{table}[!t]
    \centering
    \vspace{-4mm}
    \caption{Experimental results on the subsets of PIDray-A dataset}
    \begin{tabular}[b]{c ccc ccc}
        \toprule
            \multirow{2.5}{*}{Method} &
            \multicolumn{3}{c}{$AP_{b}$} &
            \multicolumn{3}{c}{$AP_{m}$} \\
            \cmidrule(lr){2-4} \cmidrule(lr){5-7}
            & Easy & Hard & Hide & Easy & Hard & Hide\\
        \midrule
            Mask R-CNN & 66.2&58.6&43.8&59.2&50.1&35.5\\
            Cascade MR-CNN &71.9&63.2&46.8&60.7&52.0&36.2 \\
            SDANet & 72.5&63.7&\textcolor{blue}{\textbf{48.0}}&61.1&51.7&\textcolor{blue}{\textbf{37.0}} \\
            PIDrayNet & \textcolor{blue}{\textbf{74.5}} & 64.8 & \textcolor{red}{\textbf{53.0}} & 61.4 & 51.9 & \textcolor{red}{\textbf{39.7}} \\
            RSPrompter & 74.1 & \textcolor{blue}{\textbf{67.2}} & 45.3 & \textcolor{blue}{\textbf{65.7}} & \textcolor{blue}{\textbf{58.8}} & 33.2\\
            \rowcolor{mygray}
            Our Model & \textcolor{red}{\textbf{75.7}} & \textcolor{red}{\textbf{68.3}} & 47.1& \textcolor{red}{\textbf{67.1}} & \textcolor{red}{\textbf{60.0}} & 34.3\\
        \bottomrule
    \end{tabular}
    \label{tab:subsetresults}
\end{table}%

\begin{figure}[t]
    \centering
    \includegraphics[width=0.9\linewidth]{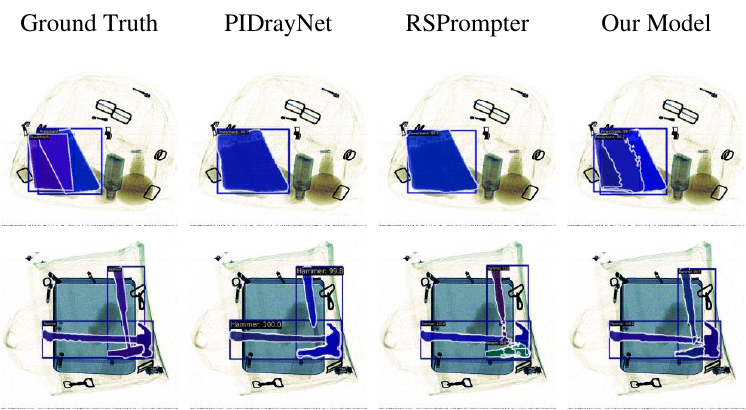}
    \caption{Qualitative results on the PIDray-A dataset to demonstrate the model's ability in segmenting occluded prohibited items.}
    \vspace{-3mm}
    \label{fig:qualitative}
\end{figure}

In addition, we provide a qualitative comparison between PIDrayNet, RSPrompter, and our proposed pipeline in Fig. \ref{fig:qualitative}. When distinguishing two overlapping power banks, our pipeline is the only model that can identify the occlusion and successfully segment the two objects separately. Additionally, ours is the only model that correctly predicts the mask of the covered hammer handle. These results indicate the effectiveness of our pipeline in distinguishing occlusions among prohibited items.

\subsection{Ablation Study}
\subsubsection{Impact on occluded prohibited items}
To demonstrate the effect of the proposed occlusion handling module, we further divide the training set of the PIDray-A into a smaller training set, the validation subset, and the occlusion subset. The validation subset contains images and annotations without occlusion, while the occlusion subset contains only occluding images and annotations. For simplicity, we discard X-ray images with no prohibited items. The approximate proportions of the split training set, validation subset, and occlusion subset are $(0.78:0.12:0.1)$.

We train our proposed pipeline, RSPrompter, and PIDrayNet on the split training set and test their performances on the validation subset and the occlusion subset to show their effectiveness in dealing with common and occluded prohibited items separately. As shown in Tab. \ref{tab:TrainValPerformance}, our proposed pipeline outperforms all other models on the validation subset. Compared to PIDrayNet and RSPrompter, our model achieves 1.5\% and 0.1\% enhancements in bounding box precision, as well as 3.2\% and 0.1\% improvements in semantic mask precision. For the occlusion subset, our proposed pipeline achieves optimal performances on 5 out of 6 metrics, highlighting the significant potential of our proposed bilayer mask decoder structure and occluder guidance operation.

\begin{table}[!t]
    \centering
    \vspace{-4mm}
    \caption{Ablation study on occluding and non-occlusion subsets}
    \begin{tabular}[b]{c | c | ccc | ccc}
        \toprule
            Set & Model & AP$_{b}$ & AP$_{b}^{50}$ & AP$_{b}^{75}$ & AP$_{m}$ & AP$_{m}^{50}$ & AP$_{m}^{75}$ \\
        \midrule
        \multirow{3}{*}{Val} & PIDNet & 77.2 & 88.1 & 84.8 & 67.3 & 87.5 & 80.0 \\
        & RSP & 78.6 & 91.9 & 88.1 & 70.6 & 91.1 & 82.8 \\
        \rowcolor{mygray}
        & Ours & \textbf{78.7} & \textbf{92.0} & \textbf{88.2} & \textbf{70.7} & \textbf{91.4} & \textbf{83.1} \\
        \midrule
        \multirow{3}{*}{Occ} & PIDNet & \textbf{49.8} & 64.0 & 55.2 & 37.8 & 60.7 & 40.3 \\
        & RSP & 46.7 & 67.7 & 54.4 & 39.4 & 63.0 & 42.5 \\
        \rowcolor{mygray}
        & Ours & 47.5 & \textbf{67.9} & \textbf{55.4} & \textbf{40.3} & \textbf{64.1} & \textbf{44.1} \\
        \bottomrule
    \end{tabular}
    \label{tab:TrainValPerformance}
\end{table}%

\begin{table}[!t]
    \centering
    \caption{Ablation study of the occluder mask decoder and residual feature guidance}
    \begin{tabular}[b]{c c | ccc | ccc}
        \toprule
            Occluder & Guide & AP$_{b}$ & AP$_{b}^{50}$ & AP$_{b}^{75}$ & AP$_{m}$ & AP$_{m}^{50}$ & AP$_{m}^{75}$ \\
        \midrule
        \XSolidBrush & \XSolidBrush & 66.1 & 83.2 & 74.2 & 56.6 & 81.3 & 64.7 \\
        \Checkmark & \XSolidBrush & 66.6 & 83.5 & 74.9 & 56.9 & 81.8 & 64.8 \\
        \Checkmark & \Checkmark & \textbf{67.5} & \textbf{84.3} & \textbf{75.8} & \textbf{57.6} & \textbf{82.6} & \textbf{65.8} \\
        \bottomrule
    \end{tabular}
    \label{tab:OccluderDecoder}
\end{table}%

\subsubsection{Impact of the occlusion-aware bilayer mask decoder module}
We further examine the impact of the occlusion-aware bilayer mask decoder module in our proposed pipeline. We first evaluate the performances with and without the residual feature addition strategy and also explore the performance when the occluder mask decoder is removed entirely, resulting in a simple SAM Mask Decoder. Tab. \ref{tab:OccluderDecoder} demonstrates the results, where the 'Occluder' column represents whether the occluder mask decoder is added, and the 'Guide' column indicates whether the residual feature guidance is applied. The results indicate that the occluder mask decoder and residual feature guidance operation together improve the overall performance of our proposed pipeline by more than 1\% on every metric, indicating the effectiveness of our proposed module.

\section{Conclusion}
In this paper, we first highlight two critical differences between prohibited items in X-ray images and natural objects: the appearance gap and the severe overlapping. To address these challenges, we present an occlusion-aware instance segmentation pipeline based on Mask R-CNN. Specifically, to generalize the priors of natural objects to prohibited X-ray items, we leverage the extensive pre-trained knowledge and strong zero-shot generalization capability of the Segment Anything Model. For handling occlusion, we propose an occlusion-aware bilayer mask decoder that explicitly models the occlusion relationships. Specifically, it uses one decoder to estimate the occlusion areas directly and adopts another standalone decoder to segment the target object guided by the estimated occlusion areas. To supervise occlusion area estimation, we introduce two large-scale X-ray datasets, PIDray-A and PIXray-A, which are annotated with additional ground-truth occlusion information. Experimental results on these datasets demonstrate the effectiveness of our pipeline.

\section*{Acknowledgments}
This research has been supported by the National Natural Science Foundation of China under Grant No. 62306258, the Major Key Project of PCL (Grant No. PCL2024A04), and the Hong Kong Polytechnic University start-up fund ZVVK.

\bibliographystyle{IEEEbib}
\bibliography{icme2025references}

\end{document}